\documentclass[twoside]{article}
\usepackage{aistats2016}

\usepackage{amsmath,amsfonts,amssymb}
\usepackage{graphicx}
\usepackage{calc}
\usepackage{color}
\usepackage{multirow}
 \usepackage{psfrag}
\usepackage{latexsym}
\usepackage{caption}
\usepackage{subcaption}
\usepackage{natbib}
\usepackage{dcolumn}
\usepackage{tikz}
\usepackage{algorithmic}
\usepackage{algorithm}
\usepackage{xspace}
\usepackage{transparent}
\usepackage{mathtools}
\usepackage{widetext}
\usepackage{units}

\newcommand{\CRM}{\mathrm{CRM}}
\newcommand{\CRP}{\mathrm{CRP}}

\newcommand{\Dir}{\mathrm{Dirichlet}}
\newcommand{\Categorical}{\mathrm{Categorical}}

\newcommand{\Poisson}{\mathrm{Poisson}}

\newcommand{\Gam}{\mathrm{Gamma}}

\newcommand{\PP}{\mathrm{PP}}

\newcommand{\zt}[1][ ]{\tilde{z}^{ #1} }

\newcommand{\tz}{ {\tilde z} }

\newcommand{\m}[1]{\boldsymbol{ #1}}
\newcommand{\mcal}[1]{\mathcal{ #1}}

\newcommand{\Levy}{\text{L\'{e}vy}\xspace}
\newcommand{\Laplace}{\text{Laplace}\xspace}

\newcommand\iidsim{\stackrel{\mathclap{\normalfont\mbox{iid} } }{\sim}}
\newcommand{\fsig}{ {\frac{1}{\sigma} } }
\newcommand{\mfsig}{ {-\frac{1}{\sigma} } }
\newcommand{\astau}{ {\alpha,\sigma,\tau} }

\newcommand{\deq}{\stackrel{d}{=}}

\newcolumntype{d}[1]{D{.}{.}{ #1} }
\usepackage{etoolbox}
\usepackage{booktabs}
\usepackage[detect-all]{siunitx}

\robustify\bfseries%

\begin{document}

\twocolumn[

\aistatstitle{Completely random measures for modelling block-structured networks
}
\aistatsauthor{
Tue Herlau \And Mikkel N. Schmidt \And Morten M{\o}rup
}
\aistatsaddress{
Technical University of Denmark \\Richard Petersens plads 31, \\ 2800 Lyngby, Denmark
\And
 Technical University of Denmark \\Richard Petersens plads 31, \\ 2800 Lyngby, Denmark
\And
 Technical University of Denmark \\Richard Petersens plads 31, \\ 2800 Lyngby, Denmark
} ]

\begin{abstract}
Many statistical methods for network data parameterize the edge-probability by attributing  latent traits to the vertices such as block structure and assume exchangeability in the sense of the Aldous-Hoover representation theorem. Empirical studies of networks indicate that many real-world networks have a power-law distribution of the vertices which in turn implies the number of edges scale slower than quadratically in the number of vertices.
These assumptions are fundamentally irreconcilable as the Aldous-Hoover theorem implies quadratic scaling of the number of edges. Recently \citet{caron2014bayesian} proposed the use of a different notion of exchangeability due to \citet{kallenberg2006probabilistic} and obtained a network model which admits power-law behaviour while retaining desirable statistical properties, however this model does not capture latent vertex traits such as block-structure.
In this work we re-introduce the use of block-structure for network models obeying Kallenberg's notion of exchangeability and thereby obtain a model which admits the inference of block-structure and edge inhomogeneity.
 We derive a simple expression for the likelihood and an efficient sampling method. The obtained model is not significantly more difficult to implement than existing approaches to block-modelling and performs well on real network datasets.
\end{abstract}

\section{Introduction}
Two phenomena are generally considered important for modelling complex network. The first is community or block structure where the vertices are partitioned into non-overlapping blocks (denoted by $\ell=1,\dots,K$ in the following) and the probability two vertices $i,j$ are connected depend on their assignment to blocks:
\begin{align}
P\big(\mbox{Edge between vertex $i$ and $j$}\big) = \xi_{\ell m} \nonumber
\end{align}
where $\xi_{\ell m} \in [0,1]$ is a number only depending on the blocks $\ell, m$ which $i, j$ belongs to. \emph{Stochastic block models} (SBMs) were first proposed by~\citet{white1976social} and today form the basic starting point for many important link-prediction methods 
 such as the \emph{infinite relational model}~\citep{xu2006learning, kemp2006learning}.

While block-structure is important for link prediction the degree distribution of edges in complex networks has also attracted a deal of attention~\citep{barabasi1999emergence,Newman2001a,strogatz2001exploring}. In many large networks the degree distribution is often found to follow a power-law~\citep{newman2010networks}
\begin{align}
P\big(\mbox{Fraction of nodes with $k$ edges}\big) & \sim k^{-\gamma}  \label{eqn:plaw}
\end{align}
where $\gamma \approx 2.5$. Explaining this scaling has lead to many important models of network \emph{growth} where edges and vertices are typically added in a sequential fashion. A notably example is the preferential attachment (PA) model of \citet{barabasi1999emergence}.

Models such as the IRM and the PA model have different goals. The PA model attempts to explain how network structure, such as the degree distribution, follows from simple rules of network growth and is not suitable for link prediction. The IRMs main goal is to discover latent block-structure and predict edges, tasks which the PA model is not suitable for. In the following we will use network model to refer to a model with the same aims as the IRM, most notably predicting missing edges.
\begin{figure*}
\centering
    \begin{subfigure}[t]{0.28\textwidth}
        \centering       %
\includegraphics[width=\linewidth]{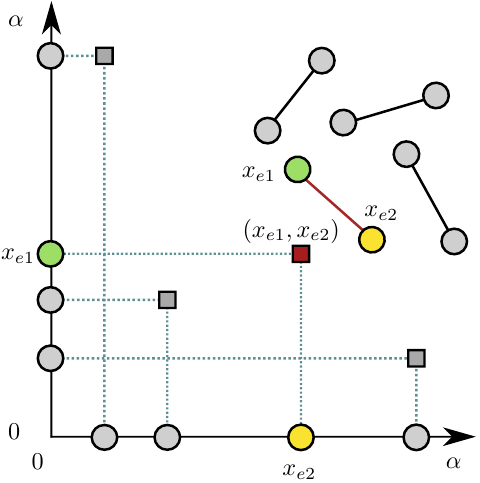}
        \caption{Maximally sparse network}\label{fig1a}
    \end{subfigure}~
    \begin{subfigure}[t]{0.36\textwidth}
        \centering   %
\includegraphics[width=\linewidth]{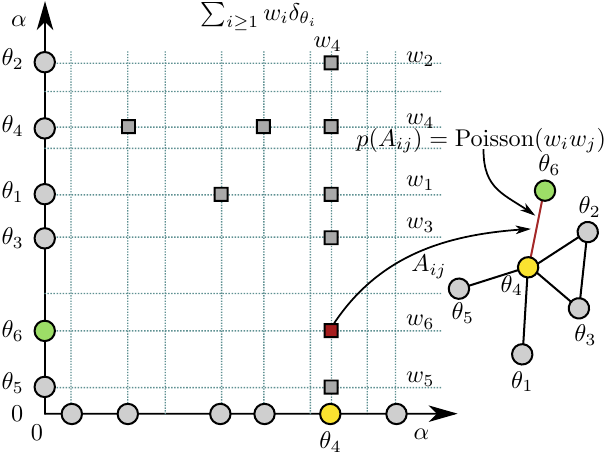}
    \caption{Nontrivial network}\label{fig1b}
    \end{subfigure}~
    \begin{subfigure}[t]{0.36\textwidth}
        \centering   %
\includegraphics[width=\linewidth]{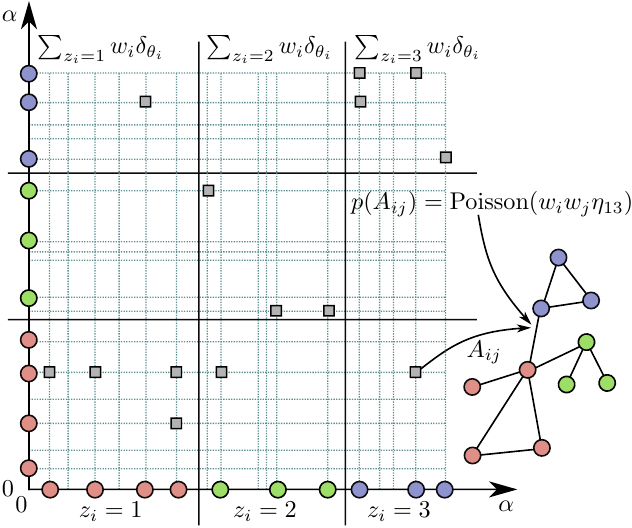}
    \caption{Nontrivial network}\label{fig1c}
    \end{subfigure}
    \caption{(Left:) A network is generated by randomly selecting points from $[0,\alpha[^2\subset\mathbb{R}_+^2$ (squares) and identifying the unique coordinates with a vertex (circles) giving the maximally disconnected graph. (Middle:) The edges are restricted to lie at the intersection of randomly generated gray lines at $\theta_i$ each with a mass/sociability parameter $w_i$ such that the probability of selecting an intersection is proportional to $w_iw_j$ giving a non-trivial network structure. (Right:) Using three independent measures $\sum_{i\geq 1} w_{\ell i}\delta_{\theta_{\ell i}}$ and modulating the interacting with a parameter $\eta_{\ell m}\geq 0$ allows block-structured networks where each random measure corresponds to a block.}\label{fig:concept}
\end{figure*}
\subsection{Exchangeability}
Invariances is an important theme in Baysian approaches to network modelling~\citep{kallenberg2006probabilistic}. For network data the criteria which has received most attention is infinite exchangeability of random arrays. Suppose we represent the network as a subset of an infinite matrix $A = (A_{ij})_{ij\geq 1}$ such that $A_{ij}$ is the number of edges between vertex $i$ and $j$ (we will allow multi and self-edges in the following). Infinite exchangeability of the random array $(A_{ij})_{ij\geq1}$ is the requirement that~\citep{hoover1979relations,aldous1981representations}
\begin{align}
(A_{ij})_{ij \geq 1} \deq (A_{\sigma(i)\sigma(j)})_{ij \geq 1} \label{eqn:infex}
\end{align}
for all finite permutations $\sigma$ of $\mathbb{N}$. The distribution of a finite network is then obtained by marginalization. An important consequence of infinite exchangeability is given by the Aldous-Hoover theorem~\citep{hoover1979relations,aldous1981representations} according to which a random network obeying infinite exchangeability eqn.~\eqref{eqn:infex} has a representation in terms of a random function and furthermore the number of edges in the network must either scale as the square of the number of vertices or (with probability 1) be zero~\citep{orbanz2015bayesian}. None of these options are compatible with a power-law degree distribution and one is faced with the dilemma of either giving up the power-law distribution or exchangeability. It is the first horn of this dilemma which have been pursued by much work on Bayesian network modelling~\citep{orbanz2015bayesian}.

It is however possible to substitute the notation of infinite exchangeability in the sense of eqn.~\eqref{eqn:infex} with a different definition due to \citet[chapter 9]{kallenberg2006probabilistic}. The new notion retains many important characteristics of the original including a powerful representation theorem parallelling the Aldous-Hoover theorem but expressed in terms of a random set. Important progress has recently been made in exploring network models based on this representation by \citet{caron2014bayesian} who demonstrate the ability to model power-law behaviour of the degree distribution and construct an efficient sampler for parameter inference. The reader is encouraged to consult this reference for more details.

In this paper, we will apply the ideas of \citet{caron2014bayesian} to block-structured network data, thereby obtaining a model based on the same structural invariance but able to capture both block-structure and degree heterogeneity.

The remainder of the paper is structured as follows. In section~\ref{section:simple} and \ref{section:measures} we will provide a basic description of the construction of \citet{caron2014bayesian} and a sketch of our proposed model, dubbed the completely random measure stochastic block model (CRMSBM). Section~\ref{section:CRM} contains a brief review of random measure theory which will be used to define the proposed model more explicitly. In section~\ref{section:sampling}  we will introduce a simple sampling scheme and section~\ref{section:experiments} will investigate the model on 11 network datasets. \section{Methods} \label{section:methods}
Before introducing the full method we will describe the construction informally omitting details relating to completely random measures.

\subsection{A simple approach to sparse networks}\label{section:simple}
Suppose the vertices in the graph are labelled by real numbers in $\mathbb{R}_+$. An edge $e$ (edges are considered directed and we allow for self-edges) then consists of two numbers $(x_{e1},x_{e2}) \in \mathbb{R}_+^2$ denoted the \emph{edge endpoint} and a network $X$ of $L$ edges (possibly $L=\infty$) is simply the collection of points $X = ((x_{e1},x_{e2}))_{e=1}^L \subset \mathbb{R}_+^2$. We adopt the convention that multi-edges implies duplicates in the list of edges.

Suppose $X$ is generated by a Poisson process with base measure $\xi$ on $\mathbb{R}_+^2$:
\begin{align}
X \sim \PP\big(\xi \big). \label{eqn:PP}
\end{align}
a \emph{finite} network $X_\alpha$ can then be obtained by considering the restriction of $X$ to $[0,\alpha[^2$: $X_\alpha = X \cap [0,\alpha[^2$. This representation easily admits sparse networks as illustrated in the left pane of figure~\ref{fig1a}. Suppose $\xi$ is the Borel measure, the number of edges is then $L \sim \Poisson(\alpha^2)$ and the edge-endpoints $x_{e1}, x_{e2}$ are 
 i.i.d. on $[0,\alpha[$. The edges are indicated by the gray squares in figure~\ref{fig1a} and the vertices as circles. Notice the vertices will be distinct with probability 1 and the procedure therefore gives rise to the
  de-generated but \emph{sparse} network of $2L$ vertices and $L$ edges shown in figure~\ref{fig1a}.

To generate non-trivial networks the edge-endpoints must co-inside with nonzero probability. Similar to \citet{caron2014bayesian}, suppose the coordinates are restricted to only take a countable number of potential values, $\theta_1,\theta_2,\dots \in \mathbb{R}_+$ and each value has an associated \emph{sociability} (or \emph{mass}) parameter $w_1,w_2,\dots \in [0,\infty[$ (we use the shorthand $(\theta_i)_i = (\theta_i)_{i=1}^\infty$ for series). If we define the measure $\mu = \sum_{i\geq 1} w_i \delta_{\theta_i}$ and let $\xi = \mu \times \mu$, then generating $X_\alpha$ according to the procedure of eqn.~\eqref{eqn:PP} the position of the edges is still selected i.i.d., but with probability proportional to $w_i w_j$ of selecting coordinate $(\theta_i,\theta_j)$. Since the edge-endpoints coincide with non-zero probability this
  procedure allows the generation of non-trivial associative network structure, see figure~\ref{fig1b}. With proper choice of $(w_i,\theta_i)_{i \geq 1}$ these networks exhibit many desirable properties such as a power-law degree distribution and sparsity~\citep{caron2014bayesian}.

This process can be intuitively extended to block-structured networks as illustrated in figure~\ref{fig1c}: Each vertice is assigned a \emph{sociability parameter} (here indicated by the colors and the symbol $z_i \in \{1,\dots,K\}$) indicating the assignment of vertex $i$ to one of $K$ blocks. 
 We can then consider a measure of the form
\begin{align}
\xi = \sum_{i,j\geq 1} \eta_{z_i z_j} w_i w_j \delta_{(\theta_i,\theta_j)} = \sum_{\ell,m=1}^K\eta_{\ell m} \mu_\ell\times \mu_m
\label{eqn:XImod}
\end{align}
Where we have introduced $\mu_\ell = \sum_{i : z_i = \ell} w_i \delta_{\theta_i}$. $\xi$ is then a measure on $[0,\alpha[^2$ and $\eta_{\ell m}$ parameterize the interaction strength between community $\ell$ and $m$. Notice the number of edges $L_{\ell m}$ between block $\ell$ and $m$ is, by basic properties of the Poisson process, distributed as $L_{\ell m} \sim \Poisson(\eta_{\ell m} T_\ell T_m)$ where $T_\ell = \mu_\ell([0,\alpha[)$ and in figure~\ref{fig1c} the location $\theta_i$ of the vertices have been artificially ordered according to color for easy visualization. The following section will show the connection between the above construction of eq.~\eqref{eqn:XImod} and the exchangeable representation due to \citet{kallenberg2006probabilistic}, however for greater generality we will let the latent trait be a general continuous parameter $u_i \in [0,1]$ and later show block-structured models can be obtained as a special case.

\subsection{Exchangeability and point-process network models}\label{section:measures}
Since the networks in the point-set representation are determined by the properties of the measure $\xi$, invariance (i.e. exchangeability) of random point-set networks is defined as invariance of this random measure. Recall infinite exchangeability for infinite matrices eqn.~\eqref{eqn:infex} was the requirement the distribution of the random matrix did not change by permutation of rows/columns in the network. For a random measure on $\mathbb{R}_+^2$ the corresponding requirement is that it should be possibly to partition $\mathbb{R}_+$ into intervals $I_1,I_2,I_3,\dots$, permute the intervals, and the random measure should be invariant to this permutation. Formally, a random measure $\xi$ on $\mathbb{R}_+^2$ is then said to be \emph{jointly exchangeable} if $\xi \circ (\varphi \otimes \varphi)^{-1} \deq \xi$ for all measure-preserving transformation  $\varphi$ of $\mathbb{R}_+$. According to \citet[theorem 9.24]{kallenberg2006probabilistic} this is ensured provided the measure has a representation of the form:
\begin{align}
\xi = \sum_{i,j\geq 1} h(\zeta,x_i,x_j) \delta_{(\theta_i,\theta_j)} \label{eqn:kallenberg}
\end{align}
where $h$ is a measurable function, $\zeta$ is a random variable and $\{(x_i,\theta_i)\}_{i\geq 1}$ is a unit rate Poisson process on $\mathbb{R}_+^2$ (the converse involves five additional terms~\citep{kallenberg2006probabilistic}). In this representation, the locations $(\theta_i)_i$ and the parameters $(x_i)_i$ are de-coupled, however we are free to select the random parameters $(x_i)_{i\geq 1}$ to lie in a more general space than $\mathbb{R}_+$.
 Specifically, define $$
 x_i = (u_i,v_i) \in [0,1] \times\mathbb{R}_+$$
 with the interpretation that each $v_i$ corresponds to a random mass $w_i$ through a transformation $w_i = g(v_i)$ and each $u_i \in [0,1]$ is a general \emph{latent trait} of the vertex (In figure~\ref{fig:concept} this parameter corresponded to the assignment to blocks). We then consider the following choice:
\begin{align}
h(\zeta,x_i,x_j) = f(u_i,u_j)g_{z_i}(v_i)g_{z_j}(v_j)
\end{align} 
where $f : [0,1]^2 \rightarrow \mathbb{R}_+$ is a measurable function (compare to the graphon in the Aldous-Hoover representation) and
$\{(u_i,v_i,\theta_i)\}_{i \geq 1}$ follows a unit-rate Poisson process on $[0,1]\times \mathbb{R}_+^2$.

\begin{figure*}
\includegraphics[width=\linewidth]{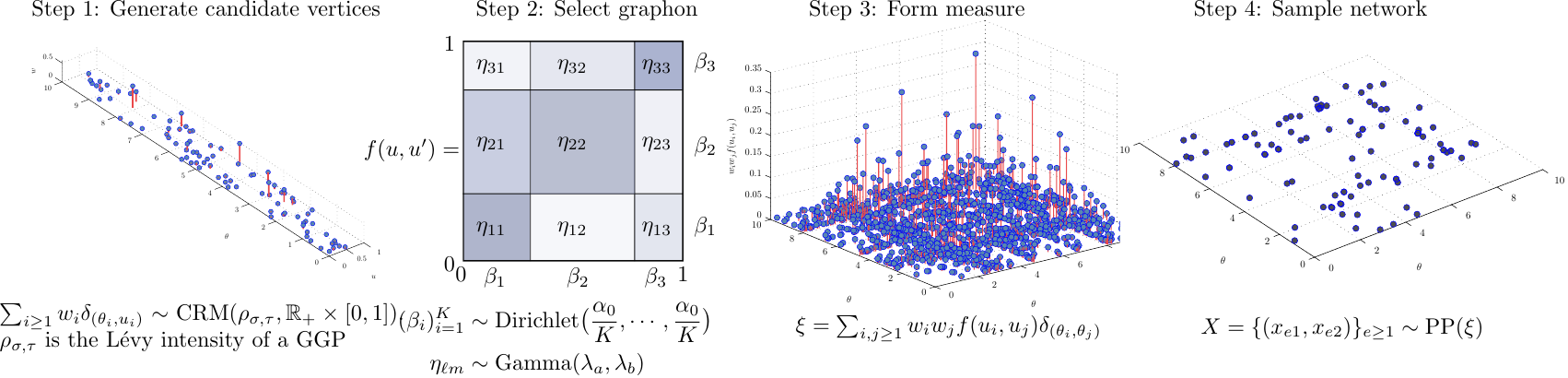}

\caption{
(Step 1:) The potential vertex locations, $\theta_i$, latent traits $u_i$ and sociability parameters $w_i$ are generated using a \emph{generalized gamma process}
(Step 2:) The interaction of the latent traits $f : [0,1]^2 \rightarrow \mathbb{R}_+$, \emph{the graphon}, is chosen to be a piece-wise constant function
(Step 3-4:) Together, these determine the random measure $\xi$ which is used to generate the network from a Poisson process
}\label{fig:concept2}
\end{figure*}

To see the connection with the block-structured model, suppose the function $f$ as the piece-wise constant function
$$f(u,u') = \sum_{\ell,m=1}^K \eta_{\ell m} 1_{J_\ell}(u)1_{J_m}(u')$$ where $$J_\ell = \left[\sum_{m=1}^{\ell-1}\beta_m, \sum_{m=1}^{\ell}\beta_m\right[, \quad \sum_{\ell=1}^K \beta_{\ell} = 1$$ and $\beta_\ell > 0$ and $z_i = \ell$ denotes the event $1_{J_\ell}(u_i) = 1$. Notice this choice for $f$ is exactly equivalent to the graphon for the block-structured network model in the Aldous-Hoover representation~\citep{orbanz2015bayesian}. The procedure is illustrated in figure~\ref{fig:concept2}.

\begin{figure}
\includegraphics[width=\linewidth]{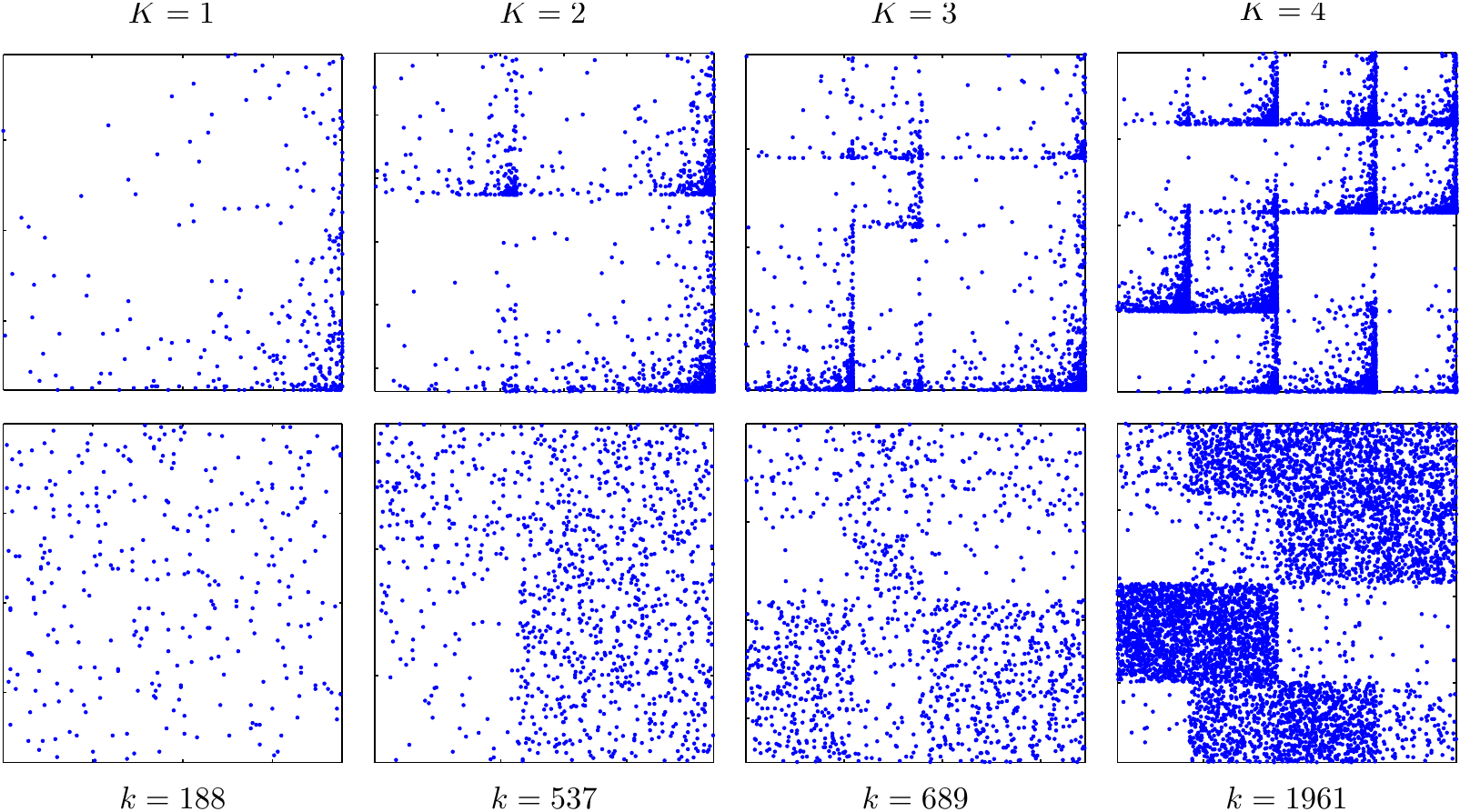}
\caption{(Top row:) Example of four randomly generated graphs for $K=1, 2, 3$ and $4$. The other parameters were fixed at $\alpha=20, \tau=1, \sigma = 0.5$ and $\lambda_a=\lambda_b=1$. Vertices has been sorted according to their assignment to blocks and sociability parameters. (Bottom row:) The same networks as above but applying a random permutation to the edges within each tile. A standard SBM assumes a network structure of this form.
}\label{fig2}
\end{figure}

 Realizations of this point-process is shown in figure~\ref{fig2} for various values of $K$ (top-row) and the choice of random measure introduced in section~\ref{section:CRM} and using the simulation method of \citet{caron2014bayesian}.  For comparison the bottom row indicates the corresponding standard stochastic block-model representation where the edges are distributed uniformly within each tile. Notice the $K=1, \eta_{11}=1$ case corresponds to the method of \citet*{caron2014bayesian}.

To fully define the method we must first introduce the relevant prior for the measure $\mu = \sum_{i \geq 1}w_i \delta_{(\theta_i,u_i)}$. As a prior we will use the Generalized Gamma-process (GGP)~\citep{hougaard1986survival}. In the following section we will briefly review properties of completely random measures and use these to derive a simple expression of the posterior.


\subsection{Random measures}\label{section:CRM}
As a prior for $\mu$ we will use completely random measures (CRMs) and the reader is referred to \citep{kallenberg2006probabilistic,kingman1967completely} for a comprehensive account. Recall first the definition of a CRM. Assume $\mathbb{S}$ is a separable complete metric space with the Borel $\sigma$-field $\mcal{B}(\mathbb{S})$ (comparing to the preceding discussion $\mathbb{S} = [0,\alpha[$). A \emph{random measure} $\mu$ is a random
variable whose values are random measure on $\mathbb{S}$. For each measurable set $A \in \mcal{B}(\mathbb{S})$, the random measure induces a random variable $\mu(A)$, and the random measure $\mu$ will be said to be \emph{completely random} if for any finite collection $A_1,\dots,A_n$ of disjoint measurable sets the random variables $\mu(A_1),\dots,\mu(A_n)$ are independent.  It was shown by \citet{kingman1967completely} any such random measure $\mu$ is discrete almost certainly with a representation
\begin{align}
\mu = \sum_{i=1}^\infty w_i \delta_{\theta_i} \label{eqn:CRM}
\end{align}
where the sequence of \emph{masses} and \emph{locations} $(w_i,\theta_i)_i$ (also known as the \emph{atoms}) is a Poisson random measure on $\mathbb{R}^+ \times \mathbb{S}$ with mean measure $\nu$. The mean measure $\nu$ thus fully characterizes the CRM and is known as the \Levy intensity measure. We will consider \emph{homogeneous} CRMs where locations are independent, $\nu(dw,d\theta) = \rho(dw)\kappa_\alpha(d\theta)$, 
 and assume $\kappa_\alpha$ is the standard Borel measure on $[0,\alpha[$. 

Since the construction as outlined in figure~\ref{fig1c} depends on sampling the edge start and end-points at random from the locations $(\theta_i)_i$ with probability proportional to $w_i$ of particular interest will be the normalized form of eqn.~\eqref{eqn:CRM}. Specifically, the chance of selecting a particular location from a random draw is governed by
\begin{align}
P = \frac{\mu}{T} = \sum_{i=1}^\infty p_i \delta_{\theta_i},\ p_i = \frac{w_i}{T}, \ T = \mu(\mathbb{S}) = \sum_{i=1}^\infty w_i \label{eqn:NRM}
\end{align}
which is known as the \emph{normalized random measure} (NRM) and $T$ is known as the \emph{total mass} of the CRM $\mu$. A random draw from a Poisson process based on the CRM can thus be realized by first sampling the number of generated points, $L \sim \Poisson(T)$, and then drawing their locations iid. from the NRM of eqn.~\eqref{eqn:NRM}.

Notice for this definition to make sense the \Levy intensity must satisfy the requirements: $\int_0^\infty \rho(dw) = \infty$ and $\int_0^\infty (1-\exp(-w))\rho(dw) < \infty$, guaranteeing respectively that $T$ is positive and finite, thereby ensuring eqn.~\eqref{eqn:NRM} is well-defined. The reader is referred to \citet{james2002poisson} for a comprehensive treatment. For many important \Levy intensities the density of the total mass does not have a simple analytical expression, however the \Laplace transform of the random variable $S$ corresponding to the total mass $T$ has the general form~\citep{james2002poisson}
\begin{align}
\mathcal{L}[S](u) = \Psi(u) & = \kappa(\mathbb{S})\int \rho(dw) (1-e^{-uw}). \label{eqn:Psi}
\end{align}
Consider a sequence of random draws made from a NRM of the form eqn.~\eqref{eqn:NRM}. Since the NRM is discrete almost surely there is a positive probability any two draws will select the same atom $\theta_i$. This induces a random partition $\Pi$ of $\mathbb{N}$ known as the \emph{exchangeable partition probability function} where $i,j \in \mathbb{N}$ are in the same block iff. they select the same atom. This partition is exchangeable and it's properties can be derived from the choice of \Levy intensity, see \citep{kingman1978representation,pitman2006combinatorial} for additional details.

In the following we will focus exclusively on the \emph{generalized gamma process} (GGP) as the choice of intensity measure\citep{james2002poisson}. The GGP is parameterized with two parameters $\sigma, \tau$ and has the functional form
\begin{align}
\rho_{\sigma,\tau}(dw) & = \frac{1}{\Gamma(1-\sigma)} w^{-1-\sigma} e^{-\tau w} dw \nonumber
\end{align}
The parameters $(\sigma,\tau)$ may lie in either the regions $]-\infty, 0]\times ]0,\infty]$ or $]0,1[\times [0,\infty[$. We will focus exclusively on the later region as the first corresponds to a \emph{finite} number of atoms. Notice this is the same choice as in \citep{caron2014bayesian}.
In conjunction with $\kappa_\alpha$ we thus obtain three parameters $(\alpha,\sigma,\tau)$ which fully describes the CRM and induced partition structure. To signify this dependence we will use $\mu_{\astau}$ for the CRM $\mu$. For this particular choice the eqn.~\eqref{eqn:Psi} becomes~\citep{pitman2006combinatorial}
 \begin{align}
  \psi_{\alpha,\sigma,\tau}(u) & = \int \rho_{\alpha,\sigma,\tau}(dw) (1-e^{-uw}) \nonumber \\
  &= \frac{\alpha}{\sigma}\left[ (u+\tau)^\sigma - \tau^\sigma \right]\label{eqn:Psimine}
  \end{align}
For $\tau=0,\alpha=1$ this is the \Laplace transform of a $\sigma$-stable random variable $S_{\sigma}$~\citep{pitman2006combinatorial,devroye2014simulation} which according to Zolotarev's integral representation has the following form~\citep{zolotarev1964representation} 
 \begin{align}
f_\sigma(x) & = \frac{\sigma x^{\frac{-1}{1-\sigma}}}{\pi(1-\sigma)} \int_0^\pi\! du\ A(\sigma,u) e^{-A(
\sigma, u)/x^{\sigma/(1-\sigma)}}\nonumber \\
A(\sigma,u) & = \left[ \frac{ \sin( (1-\sigma)u)^{1-\sigma} \sin(\sigma u)^\sigma }{ \sin(u) } \right]^{\frac{1}{1-\sigma} }.\label{eqn:sigstable} 
\end{align}
Thus introducing a new random variable $T_{\astau}$ with density
\begin{align}
g_{\alpha,\sigma,\tau}(t) & = \theta^\mfsig f_\sigma(t \theta^\mfsig) \phi_{\lambda}(t \theta^\mfsig ) \label{eqn:g}
\end{align}
where $\phi_\lambda(t) = e^{\lambda^\sigma - \lambda t}$, $\lambda = \tau \theta^\fsig$, $\theta = \frac{\alpha}{\sigma}$. Standard properties of the Laplace transform reveals $\mathcal{L}[g_\astau] = \Psi$ as defined in eqn.~\eqref{eqn:Psimine}. Thus, eqn.~\eqref{eqn:g} is the density of total mass $T$ for the GGP. Additional details, including efficient sampling methods for this distribution, is discussed by \citet{devroye2014simulation}.

With the notation in place we provide the final form of the generative process. Suppose the random measure $\mu$ (restricted to the region $[0,\alpha[$) have been generated from a GGP. Assume $z_i = \ell$ iff. $u_i \in J_\ell$ and define the $K$ sub-measures on $[0,\alpha[$:
$$
\mu_{\ell} = \sum_{i : z_i = \ell } w_i \delta_{\theta_i}
$$
each with total mass $T_{\ell} = \mu_\ell([0,\alpha[)$. The number of points in each tile $L_{\ell m}$ is then $\Poisson(\eta_{\ell m} T_\ell T_m)$ distributed, and given $L_{\ell m}$ the edge-endpoints $(x_{e1\ell},x_{e2m})$ between atoms in measure $\ell$ and $m$ can then be drawn from the corresponding NRM. The generative process is then simply:
\begin{align*}
(\beta_\ell)_{\ell=1}^K & \sim \Dir(\frac{\beta_0}{K}, \dots, \frac{\beta_0}{K}) \\
\mu & \iidsim \CRM(\rho_{\sigma,\tau}, U_{[0,1]} \times U_{\mathbb{R}_+}) \\
\eta_{\ell k} & \iidsim \Gam(\lambda_a, \lambda_b) \\
L_{\ell m} & \iidsim \Poisson(\eta_{\ell m}T_\ell T_m) \\
\mbox{for $e\!=\!1,\dots,L_{\ell m}$:}\  x_{e1\ell} & \iidsim \Categorical\big(  ( \nicefrac{ w_{i} }{ T_\ell } )_{z_i = \ell}\big) \\
\text{and } \ x_{e2m} & \iidsim \Categorical\big( \nicefrac{  w_{j} }{ T_m }) _{z_j = m}\big)
\end{align*}
  For sampling random graphs we used the same adaptive thinning strategy as \citet{caron2014bayesian}\footnote{We thank the authors for generously making their code available online.}.

\subsection{Posterior Distribution} \label{section:posterior}
In order to define a sampling procedure of the CRMSBM we must first characterize the posterior distribution. In \citet{caron2014bayesian} this was calculated using a specially tailored version of Palms formula. In this work we will use a counting argument inspired by  \citet[eqn.~(32)]{pitman2003poisson}.

To this end, consider first the case where the interaction strengths $(\eta_{\ell m})_{\ell m}$ and block sizes $(\beta_\ell)_\ell$ has a fixed value and that number of edges $L_{\ell m}$ within each tile $(\ell,m)$ is given.
Since not all potential vertices (i.e. terms $w_i \delta_{\theta_i}$ in $\mu$) will have edges attached to them it is useful to introduce a variable which encapsulates this distinction. We therefore define the variable $\tz_i = 0,1,\dots,K$ with the definition:
$$
\zt_i = \left\{\begin{array}{ll}
               z_i &\!\! \mbox{\parbox{10cm}{if there exist $(x,y) \in X_\alpha$ st. $\theta_i \in \{x,y\}$,} }  \\
               0 &\!\! \mbox{otherwise}.
             \end{array} \right.
$$
Suppose in addition for each measure $\mu_\ell$, the end-points of the edges associated with this measure selects
$$k_\ell = |\{i : \zt_i = \ell\}|$$ unique atoms and that the number of edge-endpoints selecting any particular atom $w_{i}$ is $n_{i}$. This naturally divides the edge-endpoints associated with a particular measure $\ell$ into a partition, $\{B_1,\dots,B_{k_\ell}\}$~\citep{pitman2003poisson}, and we denote by $\Pi_{\ell,2L}$ this random partition for measure $\ell$. For a particular measure the joint distribution
\begin{align}
P( \Pi_{\ell,2L} = \{B_1,\dots,B_{k_\ell}\},w_{i} \in dw_{i}, T_\ell \in dT_\ell) \label{eqn:PPart}
\end{align}
is obtained from three contributions (with $\alpha_\ell \equiv \beta_\ell \alpha$):
\begin{itemize}
\item The mass parameter $T_{\alpha_\ell}$ is distributed as $g_{\alpha_\ell,\sigma,\tau}$
\item For each $\ell=1,\dots,K$, there must be a Poisson atom in $dw_{i}$ for each $i$ such that $\zt_i = \ell$
\item For each $\ell$, we know there are Poisson atoms in $(dw_{i})_{\zt_i=\ell}$, however since the measure of these intervals is infinitesimal, the remaining mass $T_\ell - \sum_{\zt_i = \ell} w_{i}$ is still distributed as $g_{\alpha_\ell,\sigma,\tau}$.
\item Each edge-endpoint selects the atom independently with probability given by the NRM of eqn.~\eqref{eqn:NRM}, $w_{i}/T_\ell$.
\end{itemize}
The probability eqn.~\eqref{eqn:PPart} can then be obtained from these three contributions as (with $k = \sum_{\ell=1}^K k_\ell$ being the total number of vertices in the network):
\begin{align}
& \left\{
\prod_{i=1}^{k}\alpha\rho_{\sigma, \tau}(dw_{i})\right\} \\
& \times \prod_{\ell}^K \left\{
g_{\alpha_\ell,\sigma,\tau}(T_\ell-\sum_{i : \zt_i = \ell}w_{i})\right\}
\left\{
\prod_{i : \zt_i = \ell} \left( \frac{ w_{i}}{T_{\ell}} \right)^{n_{i}}
\right\} \label{eqn:PPartb}
\end{align}
where $n_{i}$ is the total number of times a particular atom $i$ of $\mu_\ell$ is selected in the process. To connect these definitions to actual network data, i.e. an array $(A_{ij})_{i,j=1}^k$, notice if the atom $(w_{i},\theta_{i})$ corresponds to a particular vertex $i$ in the network then $n_{i} = \sum_{j}(A_{ij} + A_{ji})$.

 Returning to eqn.~\eqref{eqn:PPartb} for a particular $\ell$, the expression can be integrated by introducing the variables $s_\ell = \sum_{i : \zt_i=\ell}w_{i}$ corresponding to the sum of the \emph{selected} atoms, introducing the parameters $x_{i} = w_{i}/s_{z_i}$, and integrating~\citep{pitman2003poisson,lijoi2008investigating,favaro2013mcmc}.
 With $n_\ell = \sum_{i : \zt_i = \ell} n_{i}$ eqn.~\eqref{eqn:PPartb} can be written as a product over $K$ factors:
\begin{align}
\prod_{i : \zt_i = \ell}^k\!\!(1\!-\!\sigma)_{n_i}\!\! \int_{0}^{T_\ell}\! ds_\ell \frac{ s_\ell^{n_\ell - k_\ell\sigma - 1}  g_\astau(T_\ell\!-\!s_\ell) }{\Gamma(n_\ell-k_\ell\sigma) T_\ell^{n_\ell}\alpha_\ell^{-k_\ell}  e^{\tau s_\ell }}. \label{eqn:PPartc}
\end{align}
Recall the number of edges within each tile $L_{\ell m}$ is Poisson with rate $\eta_{\ell m}T_\ell T_m$. In addition, when considering a concrete observed data matrix the edges does not have a particular labelling which is otherwise introduced in the proceeding counting argument. Thus, if we observe a number $A_{ij}$ of edges between vertices $i,j$ in a particular tile, we must consider all ways a network with this number of edges can be obtained by our generative process. This is equivalent to the number of ways of selecting the particular edge-counts of the total edge-counts within each tile. The multiplicity becomes the multinomial coefficient:
\begin{align}
{L_{\ell m} \choose (A_{ij})_{\zt_i=\ell,\zt_j=m} } & = \frac{L_{\ell m}!}{\prod_{\zt_i =\ell,\zt_j = m} A_{ij}!}. \label{eqn:multinom}
\end{align}
The probability of obtaining a particular observed network $A_{ij}$ can be obtained by combining eqs.~\eqref{eqn:PPartc}, \eqref{eqn:multinom} and the Poisson rates for the edge-counts within each tile to obtain:
\begin{align}
 & P(A,(z_i)_i | (\eta_{\ell m})_{\ell m}, (\beta_{\ell})_\ell ) = \left\{\prod_{\ell=1}^K\int_{0}^\infty dT_\ell \left\{ \mbox{eqn.~\eqref{eqn:PPartc}} \right\} \right\}\nonumber \nonumber \\
& \times \prod_{\ell m} \left\{ \Poisson(L_\ell | \eta_{\ell m}T_\ell T_m) \frac{L_{\ell m}!}{\prod_{\substack{ \zt_i =\ell,\\ \zt_j = m}} A_{ij}!} \right\}.
\end{align}
Defining $n_{\ell m} = \sum_{\zt_i = \ell,\zt_m=j}A_{ij}$ and simplifying
 \begin{align*}
& P(A, (z_i)_i | (\eta_{\ell m}),(\beta_{\ell})_\ell) = \frac{1}{\prod_{ij} A_{ij}!}\prod_\ell \left[\int_0^\infty\!\!\! \int_{0}^{T_\ell}\!\!dT_\ell  ds_\ell\right] \\
& \times \left[ \prod_{\ell m} \eta_{\ell m}^{n_{\ell m} }  e^{ -\sum_{\ell m} \eta_{\ell m}T_\ell T_m} \right] \left\{\prod_{\ell} E_\ell \right\} \nonumber
\end{align*}
where we have defined
$$E_\ell =
  \frac{\alpha^{k_\ell} s_\ell^{n_\ell - k_\ell\sigma - 1}
  }{\Gamma(n_\ell-k_\ell\sigma) e^{\tau s_\ell } }  g_{\alpha_\ell,\tau,\sigma}(T_\ell\!-\!s_\ell) \prod_{\zt_i=\ell} (1-\sigma)_{n_{i}}.$$
Similar to \citet{lijoi2008investigating} we will use the simple change-of-variable from $T$ to $t=T-s$ and a change in the order of integration to obtain:
 \begin{align}
\int_{\mathbb{R}^+} \int_0^s\! dTds\ h(T,s) = \iint_{\mathbb{R}_+^2} dsdt\ h(t+s,s).\label{eqn:integratestric}
\end{align}
Then introducing the Gamma-priors for $\eta_{\ell m}$, Dirichlet prior for $(\beta_\ell)_\ell$ and integrating over $\eta_{\ell m}$ we obtain the final expression:
 \begin{align}
& P(A, (z_i)_i,\sigma,\tau, (\alpha_\ell,s_\ell,t_\ell)_\ell ) = \frac{\Gamma(\beta_0) \prod_{\ell =1}^K \alpha_\ell^{\frac{\beta_0}{K} - 1} }{\Gamma(\frac{\beta_0}{K})^K \alpha^{\beta_0}} \nonumber \\
& \times \frac{\prod_{\ell} E_\ell}{\prod_{ij} A_{ij}!}
 \prod_{\ell m} \frac{G(\lambda_a\!+\!n_{\ell m}, \lambda_b\!+\!(s_\ell\!+\!t_\ell)(s_m\!+\!t_m))}{G(\lambda_a, \lambda_b)}
 \label{eqn:logp}
\end{align}
where $G(a,b) = \Gamma(a)b^{-a}$ is the normalization factor of the Gamma distribution. Finally notice the $\eta=1$ case, corresponding to \citet{caron2014bayesian}, can be obtained by taking the limit $\lambda_a=\lambda_b\rightarrow \infty$ in which case
$\frac{G(\lambda_a + n,\lambda_b + T)}{G(\lambda_a, \lambda_b)} \rightarrow e^{-T}$.
When discussing the $K=1$ case we will assume this limit has been taken.
\subsection{Inference}\label{section:sampling}
Sampling the expression eqn.~\eqref{eqn:logp} requires three types of sampling updates: For $A_{ij}$ we must apply a sampling procedure to impute missing values, the sequence of block-assignments $(z_i)_i$ must be updated, the parameters associated with the random measure $\sigma, \tau$ must be updated and finally the remaining variables $(\alpha_\ell,s_\ell,t_\ell)$ associated with each expression $E_\ell$ must be updated. We will first consider the later problem:

\paragraph{Update of variables associated with each $E_\ell$:}
 All terms except the densities $g_\astau$ are amenable to standard sampling techniques. In \citep{caron2014bayesian} this expression was sampled by employing a proposal distribution proportional to the density, thus allowing their value to cancel. In our work we opted for the approach of \citet{lomeli2014marginal} in which $u$ in Zolotarev's integral representation eqn.~\eqref{eqn:sigstable} is considered an auxiliary parameter. Thus, introducing
 $u_\ell \in ]0,\pi[$ for each RPM gives the full set of variables $\Phi_\ell = (\alpha_\ell, s_\ell, t_\ell,u_\ell)$ for each RPM. For convenience, the domain of the variables are in turn transformed to $\mathbb{R}$ using the standard change-of-variables $x \mapsto e^{x}$ for $\alpha,t$ and $s$ and the logistic mappings $x \mapsto (1+e^{-x})^{-1}$, $x \mapsto \pi(1+e^{-x})^{-1}$ for $\sigma$ and $u$. We found a simple 
 random-walk Metropolis-Hastings sampling with a $\mathcal{N}(0,\sigma=0.1)$ kernel (50 steps per iteration) was robust and efficient compared to the other updates. 
\paragraph{Update of $z_i$:}
These variables can be updated directly from the likelihood eqn.~\eqref{eqn:logp}, however we opted to re-impute the weights $(w_i)_{\zt_i = \ell}$ by
 inverting the integration step from eqn.~\eqref{eqn:PPartb} to eqn.~\eqref{eqn:PPartc} to obtain
\begin{align}
\m (w_{i}/s_\ell)_{i : \zt_i = \ell} & \sim \Dir\left( (n_{i}-\sigma)_{i : \zt_i = \ell} \right)
\label{eqn:wimpute}
\end{align}
doing this for each $\ell=1,\dots,K$ allows all variables $z_i$ to be updated in a regular Gibbs sweep.
\paragraph{Update of $A_{ij}$:}
Most networks are binary whereas the model assumes count-data. Furthermore to test the model it is useful to predict the presence of unobserved edges. Both of these difficulties are resolved by imputation. Suppose we are given a matrix $W$ such that $W_{ij} = 1$ iff. the edge-count $A_{ij}$ is unobserved. Furthermore assume $A_{ij}$ is binary and must be imputed. Edges can then in principle be imputed directly by performing MCMC updates of $A_{ij}$ and accepting/rejecting according to the likelihood eqn.~\eqref{eqn:logp}, however the coupling between different counts through the gamma functions in $E_\ell$ would make such a sampling procedure prohibitively expensive. This difficulty is not present in \citet{caron2014bayesian} where the sociability-vector $(w_{i})_i$ are retained and updates using Hamiltonian Monte-Carlo, however we can re-sample $(w_{i})_i$ and $(\eta_{\ell m})_{\ell m}$ from their marginal distributions and use the re-sampled values of $(w_{i})_i$ to impute the corresponding values of $(A_{ij})$. Thus for each plate $(\ell,m)$ we sample $(w_i)_{\zt_i = \ell}$ from \eqref{eqn:wimpute} and $\eta_{\ell m}$ from
\begin{align}
\eta_{\ell m} \sim \Gam\big(n_{\ell m}\!+\!\lambda_a, (t_\ell\!+\!s_\ell)(t_m\!+\!s_m)\!+\!\lambda_b\big)
\label{eqn:etaimpute}
\end{align}
the distribution of each unobserved $A_{ij}$ is then simply $\Poisson(\eta_\ell w_{i} w_{j})$, $z_i=\ell, z_j=m$.
\begin{figure}
\includegraphics[width=\linewidth]{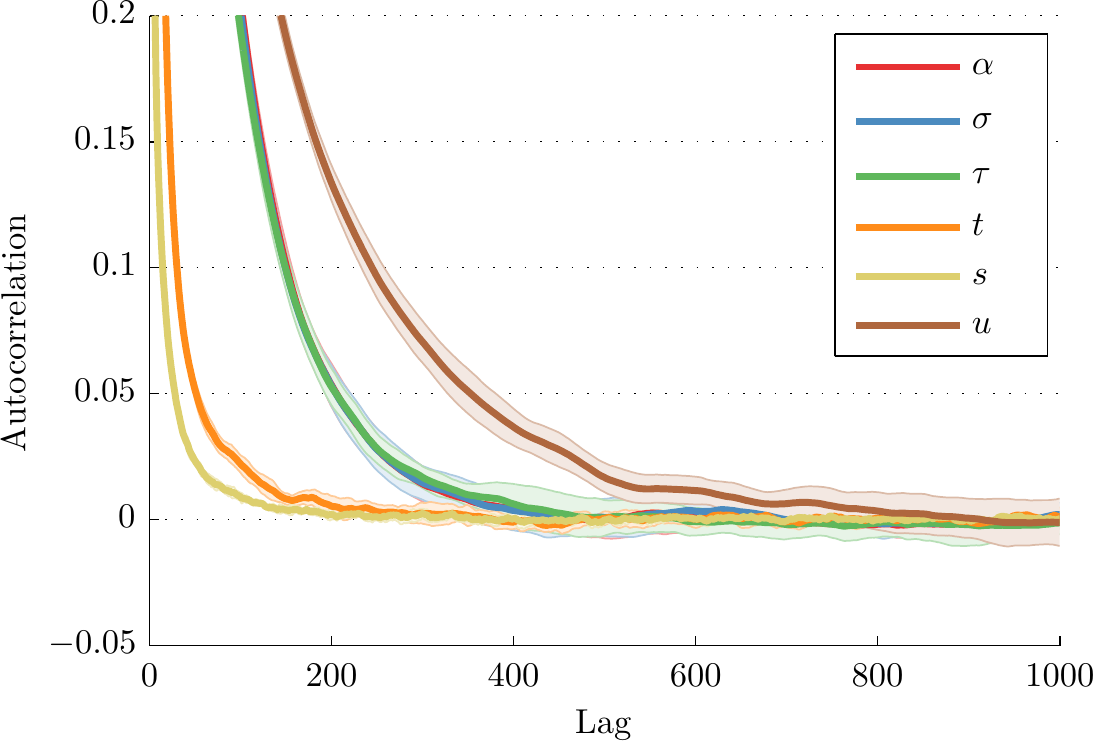}
\caption{Autocorrelation plots for lags up to 1\,000 of the parameters $\alpha,\sigma,\tau$ (top row) and $s,t,u$ (bottom row) for a $K=1$ network drawn from the prior distribution using $\alpha=25$, $\sigma=0.5$ and $\tau=2$. The plots were obtained by evaluating the proposed sampling procedure for $10^6$ iterations and the shaded region indicates standard deviation obtained over 12 re-runs. The simulation indicates reasonable mixing for all parameters with $u$ being the most affected by excursions.}\label{fig3}
\end{figure}

\subsubsection{Full inference procedure}\label{sec:samp}
Collecting the preceding steps we obtain
\begin{itemize}
\item For $\ell=1,\dots,K$, update the four variables in $\Phi_\ell$ and $\sigma,\tau$ using random-walk metropolis hastings
\item Impute $(w_{i})_{\zt_i = \ell}$ using eqs.~\eqref{eqn:etaimpute} once for each $\ell$ and then iterate over $i$ and update each $z_i$ using a Gibbs sweep from the likelihood. 
\item Impute $(\eta_{\ell m})_{\ell m}$ and $(w_{i})_i$ using eqs.~\eqref{eqn:etaimpute} and \eqref{eqn:wimpute} and for each $(ij)$ such that the edge is either unobserved ($W_{ij} = 1$) or must be imputed ($A_{ij} \geq 1$) generate a candidate $a \sim \Poisson(\eta_{\ell m} w_{i} w_{j})$. Then, if $W_{ij}=1$ simply set $A_{ij} = a$, otherwise if $W_{ij}=0$ and $a=0$ reject the update.
\end{itemize}

The parameters $\Phi_\ell$ and $\sigma,\tau$ are important for determining the sparsity and power-law properties of the network model~\citep{caron2014bayesian} and to investigate the sampling of these parameters we generated a single network problem using $\alpha=25, \sigma=0.5, \tau=2$ and evaluated 12 samplers with $K=1$ on the problem. Autocorrelation plots (mean and standard deviation computed over 12 restarts) can be seen in figure~\ref{fig3}. All parameters mix, however the different parameters have different mixing times with in particular $u$ being affected by excursions. This indicates many slice-sampling update of $\Phi_\ell$ are required to explore the state space appreciably and we therefore applied 150 slice-sampling updates of $\Phi_\ell$ for each update of $(z_i)_i$ and $A_{ij}$. Basic validation of the sampling procedure can be found in the supplementary material.
\section{Experiments} \label{section:experiments}
The proposed method was evaluated on 11 network datasets (a description of how the datasets were obtained and prepared can be found in the supplementary material). As a criteria of evaluation we decided for AUC score on held-out edges, i.e. predicting the presence or absence of unobserved edges using the imputation method described in the previous section. All networks were initially processed by thresholds at 0 and vertices with zero edges were removed. A fraction of 5\% of the edges were removed and considered as held-out data.

\begin{figure}
\includegraphics[width=\linewidth]{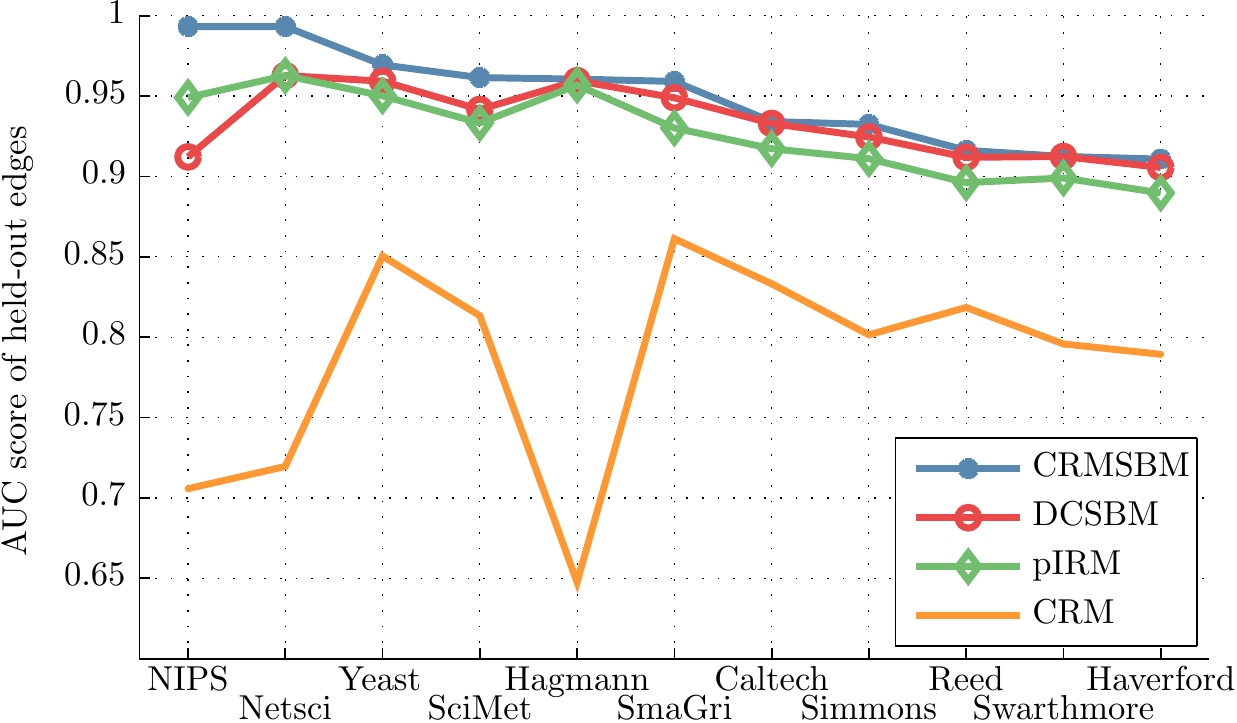}
\caption{AUC score on held-out edges for the selected methods (averaged over 4 restarts) on 11 network datasets. For the same number of blocks, the CRMSBM offers good link-prediction performance compared to the method of \citet{caron2014bayesian} (CRM), a SBM with Poisson observations (pIRM) and the degree-corrected SBM (DCSBM)~\citep{TUE}. Additional information is found in the supplementary material.}\label{fig4}
\end{figure}

To examine the effect of using blocks, we compared the method against the method of \citet{caron2014bayesian} (CRM) (corresponding to $\eta_{\ell m}=1$ and $K=1$), a standard block-structured model with Poisson observations (pIRM)~\citep{kemp2006learning} and the degree-corrected stochastic block model (DCSBM)~\cite{TUE}, a simple model which allows both block-structure and degree-heterogeneity but is not exchangeable. 
 More details on the simulations and methods are found in the supplementary material. 

The pIRM was selected since it is the closest block-structured model to the CRMSBM without degree-correction. This allows us to determine the relative benefit of inferring the degree-distribution compared to only the block-structure. 

For priors we selected uniform priors for $\sigma,\tau,\alpha$ and a $\Gam(2,1)$ prior for $\beta_0, \lambda_a,\lambda_b$. For consistency similar choices were made for the alternative models. 

All methods were evaluated for $T=2\,000$ iterations and the later half of the chains was used for link prediction. 
We used $4$ random selections of held-out edges per network to obtain the results seen in figure~\ref{fig4} (same sets of held-out edges were used for all methods). It is evident the use of block-structure is crucial to obtain good link prediction performance. For the block-structured methods, the results indicate additional benefits from using models which allows the modelling of degree-heterogenity for most networks except the Hagmann brain connectivity graph. This result is possibly explained by the Hagmann graph being constructed by fixing the number of outgoing edges for each vertex. Comparing the CRMSBM and the DCSBM, these models perform either on par or with a slight advantage to the CRMSBM.
Models of networks based on the CRM representation of \citet{kallenberg2006probabilistic} offers one of the most important new ideas in statistical modelling of networks in recent years and to our knowledge \citet*{caron2014bayesian} were the first to realize the benefits of this modelling approach as well as describing it's statistical properties and provide an efficient sampling procedure.

The degree distribution of a network is only one of several important characteristics of a complex network. In this work we have examined how the ideas presented in \citet*{caron2014bayesian} can be applied for a simple block-structured network model to obtain a model which admits block structure and degree correction. Our approach is a fairly straightforward generalization of the methods of \citet*{caron2014bayesian}. However, we have opted to explicitly represent the density of the total mass $g_{\alpha_\ell,\sigma,\tau}$ using Zolotarev's integral representation and integrate out the sociability parameters $(w_{i})_i$ reducing the number of parameters associated with the CRM from the order of vertices to the order of blocks. 

The resulting model has the increased flexibility of being able to control the degree distribution within each block. In practice results of the model on 11 real-world datasets indicates this flexibility offers benefits over a purely block-structured approaches to link prediction for most networks as well as some potential benefit over an alternative approach to modelling block-structure and degree-heterogeneity. The results heavily indicates structural assumptions (such as block-structure) is important to obtain reasonable link prediction.


Block-structured network modelling is in turn the simplest structural assumption for block-modelling. The extension of the method of  \citet*{caron2014bayesian} to overlapping blocks, possibly using the correlated random measures of \citet{chen2013dependent}, appears fairly straight-forward and should potentially offer a generalization of overlapping block models. 


\subsubsection*{Acknowledgments}
This project was funded by the Lundbeck Foundation (grant nr. R105-9813).
\newpage
\appendix
\section{Supplementary Material}
%
\subsection*{Validation of the sampler}
\begin{figure}\centering
\includegraphics[width=.8\linewidth]{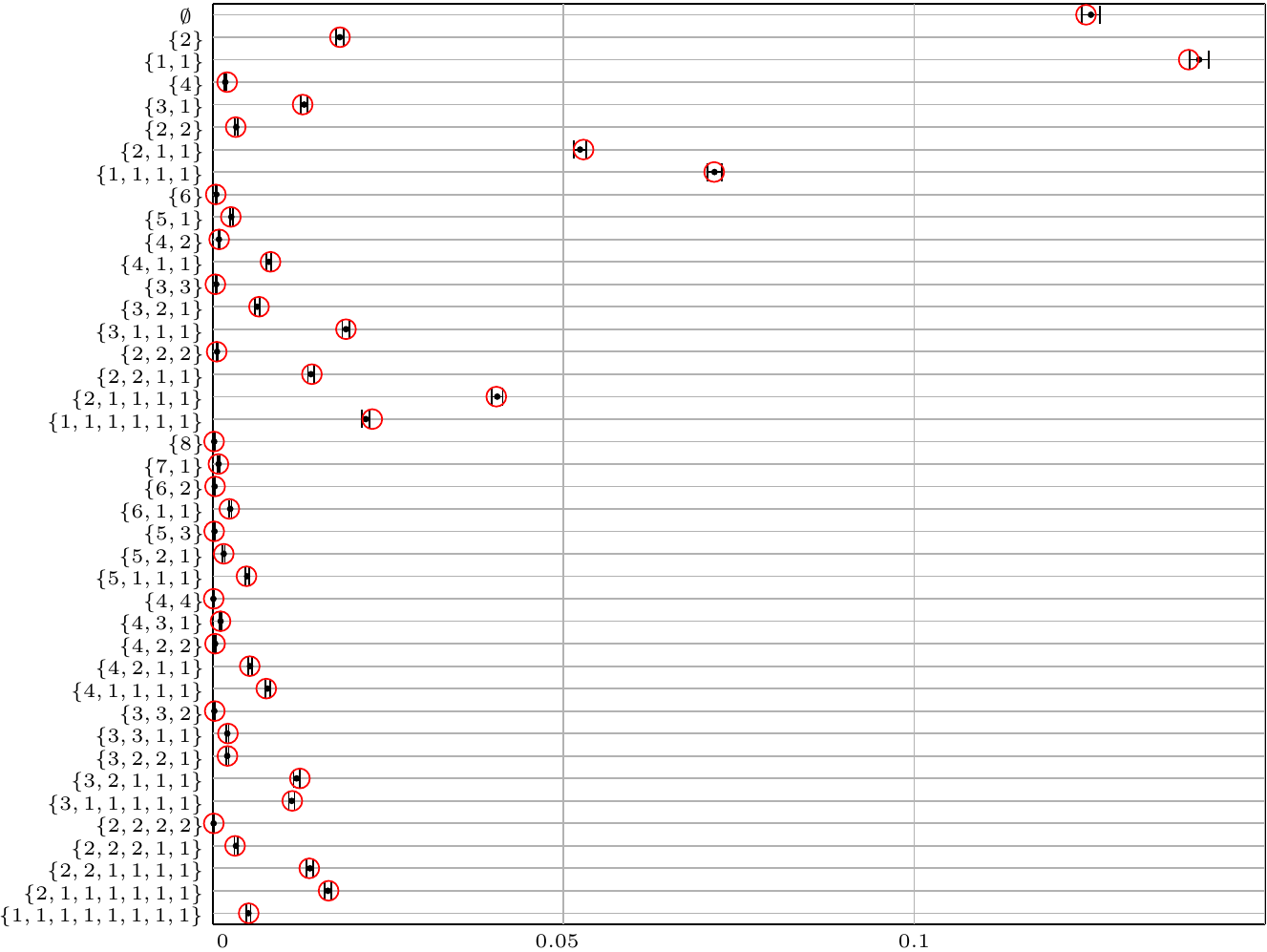}
\caption{The estimated frequency of all unique networks binned according to their unique edge-endpoint counts $(n_1,\dots,n_k)$. (ordered decreasingly) for $L=0,\dots,4$ edges (41 in total, red circles), as well as the frequency obtained by computing the probability (see text for details). }\label{fig3a}
\end{figure}

To investigate the validity of the sampling procedure, we considered the $K=1,\lambda_a=\lambda_b\rightarrow \infty$ case and used the sampling procedure of \citep{caron2014bayesian} with $(\alpha=2,\sigma=0.5,\tau=1)$ to generate $250\,000$ random networks. As described in the previous section the probability of any given network is fully determined by the edge-endpoint counts $(n_1,\dots,n_k)$ and the probability of a particular sequence of counts is permutation invariant. If ordered decreasingly this gives 41 unique vectors of edge-endpoint counts $(n_1, \dots, n_k)$ for $L=0,1,2,3,4$ (see vertical axis on figure~\ref{fig3a}) and the generated networks were binned according to their edge-endpoint count signature (networks with more than 4 edges were discarded). In this manner we obtained an estimate of the true frequency of a particular network signature.
This estimate of the frequency was compared against the probability of a given network as computed by eqn.~(17). Notice that due to permutation invariance the probability of each network signature must be corrected by multiplying eqn.~(17) with a factor obtained by a combinatorial argument (see for instance \citet[eqn.~(2.2)]{pitman2006combinatorial})
\begin{align}
\frac{n!}{\sum_{i=1}^{m_i} (i!)^{m_i} m_i! }, \quad
\text{ where } \quad m_i = \sum_{i=1}^k 1(n_i=1). \nonumber
\end{align}
We thus obtain two estimates of the probability of a particular network signature shown in figure~\ref{fig3a}, both in close agreement.
\begin{figure}\centering
\includegraphics[width=.8\linewidth]{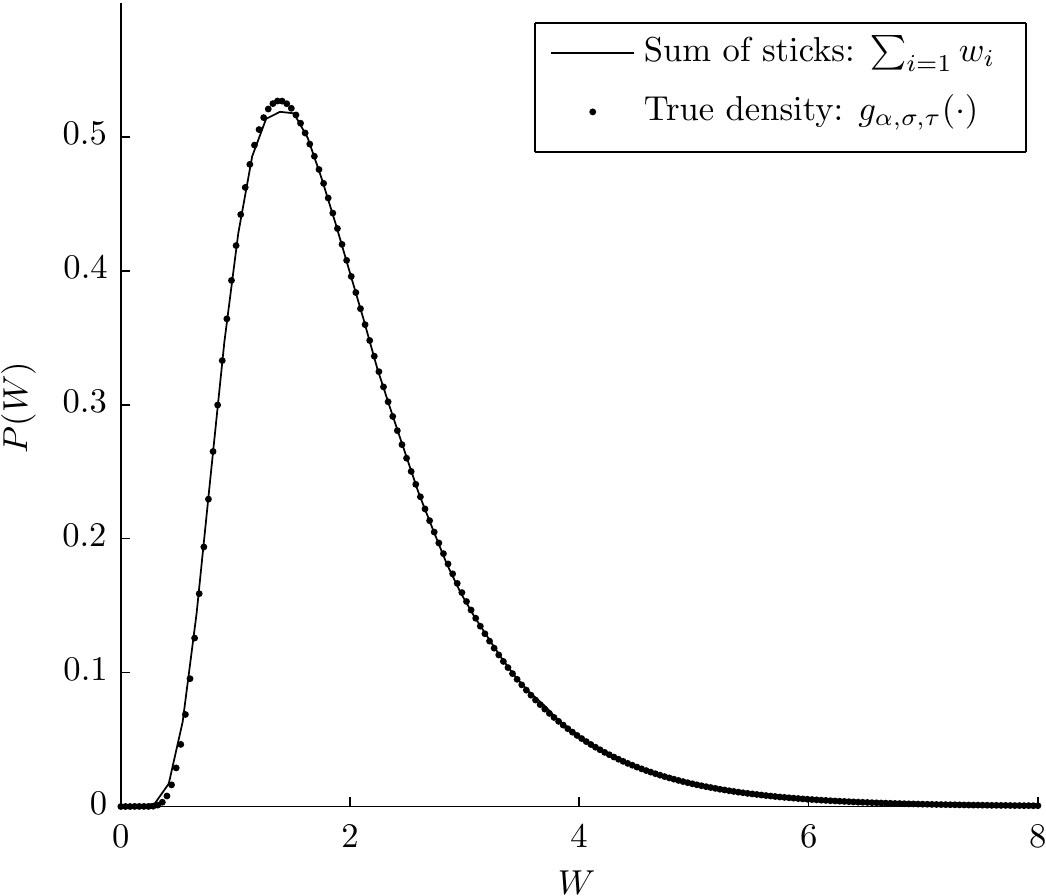}
\caption{The density of the stick length $\sum_i w_i$ for the randomly generated networks as well as the true density eqn.~\eqref{eqn:gappendix} obtained by numerical integration of eqn.~\eqref{eqn:zolo}}\label{fig3b}
\end{figure}
 In figure~\ref{fig3b} is shown the estimated density of the total mass $T$ obtained by numerically integrating Zolotarev's integral representation of $f_\sigma$
 \begin{align}
f_\sigma(x) & = \frac{\sigma x^{\frac{-1}{1-\sigma}}}{\pi(1-\sigma)} \int_0^\pi\! du\ A(\sigma,u) e^{-A(
\sigma, u)/x^{\sigma/(1-\sigma)}}, \nonumber \\
A(\sigma,u) & = \left[ \frac{ \sin( (1-\sigma)u)^{1-\sigma} \sin(\sigma u)^\sigma }{ \sin(u) } \right]^{\frac{1}{1-\sigma} }, \label{eqn:zolo} \\   
g_{\alpha,\sigma,\tau}(t) & = \theta^\mfsig f_\sigma(t \theta^\mfsig) \phi_{\lambda}(t \theta^\mfsig ) \label{eqn:gappendix}
\end{align}
and the estimated density of the total mass $T$ obtained by summing the generated sticks $(w_i)$.  Both the estimates of the networks signatures and the density of $T$ are in close agreement.

\paragraph{Datasets and preparation}
To test the methods we selected 11 publicly available datasets describing social networks, co-authorship networks and biological networks.
\begin{description}
\item[Yeast:] Interaction network of $2361$ proteins in yeast~\citep{bu2003topological}.
\item[SmaGri:] Coauthorship network of $1059$ authors from the  Garfield's collection of citation networks~\citep{batagelj2014pajek}.
\item[SciMet:] Coauthorship network of $3084$ authors from the Scientometrics journal, 1978-2000~\citep{batagelj2014pajek}.
\item[Netscience:] Coauthorship network of $1589$ authors working in network theory as compiled by ~\citet{newman2006finding}.
\item[Hagman:] Structural brain networks where edges correspond to the number of fiber tracts between $998$ brain regions. All five networks in the dataset were simply averaged to produce a single network~\citep{hagmann2008}.
\item[NIPS:] Consisting of the $2865$ authors who have coauthored papers together at the 1-12'th NIPS conference~\citep{nipsnet}.
\item[Caltech, Simmons, Reed, Haverford, Swarthmore:] Five social networks of $769, 1446, 962, 1518, 1659$ students respectively obtained from the Facebook100 dataset~\citep{facebook100}.
 \end{description}
The datasets were processes similarly by first removing any vertices without edges, i.e. where $n_i=0$, and thresholding at 0 to produce binary networks. Selection of the missing edges for link prediction was done by first removing a fraction of $5\%$ of all potential edges at random and then, if this procedure left any vertices without attached edges, re-introducing one of the edges attached to each such vertex and removing (at random from all other potential edges) a single edge. This procedure was repeated until $5\%$ of the potential edges were removed and all vertices had at least one edge attached.

\paragraph{Models considered}
In addition to non-parametric extensions of the Poisson SBM we compared the CRMSBM against a degree corrected block model, the \emph{degree-corrected stochastic block model} (DCSBM) of \citet{TUE}. This model is not exchangeable but does model block structure and sociability.

Specifically the DCSBM assumes a generative process of the form:
\begin{align*}
(z_1,\dots,z_n) & \sim \CRP(\alpha) \\
\eta_{\ell m} & \sim \Gam(\lambda_a,\lambda_b) \\
(\theta_{i\ell}^{(1)}), (\theta_{i\ell}^{(2)}) & \sim \Dir((\gamma)_{i=1}^{k_\ell}) \\
A_{ij} & \sim \Poisson(k_{z_i}k_{z_j}\theta_{iz_i}^{(1)}\theta_{jz_j}^{(2)} \eta_{z_i z_j}).
\end{align*}
To be consistent with the CRMSBM we selected a prior of the form $\Gam(2,1)$ for $\alpha, \lambda_a$ and $\lambda_b$. The model is somewhat sensitive to the choice of prior for $\gamma$ however we found a prior of the form $\Gam(2,1)$ to perform reasonably well. The DCSBM reduces to a model without degree-correction, the pIRM~\citep{kemp2006learning}, by the choice $\gamma_{i \ell} = \frac{1}{n_\ell}$.

%
\bibliographystyle{plainnat}
\bibliography{library}

\end{document}